\documentclass[final,5p,times,twocolumn]{elsarticle}

\usepackage{amssymb}
\usepackage{algorithm}
\usepackage{algorithmic}
\usepackage{amsmath}
\usepackage{amsfonts}
\usepackage{multirow}
\usepackage{color}
\usepackage{booktabs}

\DeclareMathOperator*{\argmin}{arg\,min}

\usepackage{subcaption}
\usepackage{rotating}

\newcounter{globaltheo}
\newtheorem{globalthm}[globaltheo]{Proposition} 
\newtheorem{definition}{Definition}

\usepackage{lineno}

\journal{Information Fusion}

\begin{document}

\begin{frontmatter}



\title{Discovering Common Information in Multi-view Data}



\author[label1]{Qi Zhang}
\author[label1]{Mingfei Lu}
\author[label2,label3]{Shujian Yu\corref{cor1}}
\author[label1]{Jingmin Xin}
\author[label1]{Badong Chen\corref{cor1}}
\cortext[cor1]{To whom correspondence should be addressed (yusj9011@gmail.com; chenbd@mail.xjtu.edu.cn).}
\affiliation[label1]{organization={National Key Laboratory of Human-Machine Hybrid Augmented Intelligence, National Engineering Research Center for Visual Information and Applications, Institute of Artificial Intelligence and Robotics},
            addressline={Xi'an Jiaotong University}, 
            city={Xi'an},
            postcode={710049}, 
            state={Shaanxi},
            country={China}      } 
\affiliation[label2]{organization={UiT - The Arctic University of Norway},
            city={Troms{\o}},
            postcode={9037}, 
            country={Norway}}

\affiliation[label3]{organization={Vrije Universiteit Amsterdam},
            city={HV Amsterdam},
            postcode={1081}, 
            country={The Netherlands}}
            
\begin{abstract}
We introduce an innovative and mathematically rigorous definition for computing common information from multi-view data, drawing inspiration from G\'acs-K\"orner common information in information theory.
Leveraging this definition, we develop a novel supervised multi-view learning framework to capture both common and unique information.
By explicitly minimizing a total correlation term, the extracted common information and the unique information from each view are forced to be independent of each other, which, in turn, theoretically guarantees the effectiveness of our framework. 
To estimate information-theoretic quantities, our framework employs matrix-based R{\'e}nyi's $\alpha$-order entropy functional, which forgoes the need for variational approximation and distributional estimation in high-dimensional space.
Theoretical proof is provided that our framework can faithfully discover both common and unique information from multi-view data.
Experiments on synthetic and seven benchmark real-world datasets demonstrate the superior performance of our proposed framework over state-of-the-art approaches.
\end{abstract}




\end{frontmatter}


\section{Introduction}
The advent of diverse and heterogeneous data due to recent technological advancements has spurred increasing interest in multi-view learning~\cite{zheng2021deep, lee2021variational, zheng2023comprehensive}.
This field relies on two principles: the \emph{consensus} principle, which seeks consensus information across different views, and the \emph{complementary} principle, which recognizes the unique, valuable information each view offers~\cite{wu2019multi, fu2021red, wan2021multi}.  
For instance, consider the case of an animal's binocular vision. 
Each eye captures a different yet highly correlated perspective of an object, extracting consensus information and demonstrating the consensus principle. 
Simultaneously, the differing wide-angle views of each eye provide complementary information, expanding the field of view and increasing perception accuracy, illustrating the complementary principle.

Within the domain of multi-view learning, a plethora of methods have been proposed. 
\textcolor{black}{This encompasses strategies rooted in subspace learning \cite{cai2023seeking}, approaches leveraging non-negative matrix factorization \cite{cui2022nonredundancy}, and methodologies employing multiple similarity graphs \cite{zhang2022incomplete}, among others.}
For the scope of this discourse, we will concentrate on three prevalent approaches that follow these two principles.
Firstly, canonical correlation analysis (CCA)~\cite{chaudhuri2009multi} and its derivatives \cite{feng2018angle, zhou2016linked, lofstedt2011onpls}, including \textcolor{black}{deep} CCA \cite{andrew2013deep}, \textcolor{black}{kernel} CCA~\cite{arora2012kernel} and D-GCCA \cite{shu2022d}, are representative methods based on the consensus principle in multi-view learning. 
These techniques project multiple views into a shared low-dimensional subspace, maximizing their correlation to learn consensus. 
Variations methods of this category address nonlinear and complex relationships between views.
Another grounded in the consensus principle method draws on mutual information in information theory. 
The authors in \cite{federici2020learning} posit that each view contains identical task-relevant information, a classic hypothesis suggesting that effective representation models view-invariant factors. 
They develop robust representations by maximizing the mutual information between representations from different views. 
A similar approach is used in \cite{bachman2019learning}, where information about high-level factors that span across multiple views is captured by maximizing the mutual information between the extracted features.
Recent studies \cite{wan2021multi, xu2021multi} highlight the utilization of specifically designed neural network architectures to extract both consensus and complementary information from multi-view data. 
More precisely, these studies apply variational autoencoder and variational inference techniques to learn compact representations and scrutinize consensus and complementarity across different viewpoints.
These approaches emphasize the importance of the network structure itself.    

Regardless of their advancements, each approach bears its own limitations.
The majority of CCA-based methods capture the correlation between views, rather than exact consensus information~\cite{zheng2023comprehensive, han2023trusted}.
Mutual information lacks a clear interpretation in terms of decomposing random variables into unique and common components~\cite{kleinman2022gacs}.
Although the approach that relies on the network structure itself to extract consensus and complementary information is both intuitive and engineering-focused, it may be deficient in rigorous theoretical support.
Furthermore, none of the aforementioned methods provide a clear and rigorous mathematical definition for handling consensus information in multi-view data. 

In our research, we adhere to the \emph{consensus} principle by discovering \emph{common} information from multi-view data, and uphold the \emph{complementary} principle by discerning \emph{unique} information intrinsic to each view, also referred to as view-specific~\cite{wan2021multi} or view-peculiar~\cite{xu2021multi} information.     
The task of discerning common information from multi-view data and condensing it into a singular variable continues to be a significant challenge, notably in the field of high-dimensional data analysis. 
A formal mathematical definition of common information in multi-view neural network learning literature is scarce, and no existing literature has offered a rigorous way to extract such kind of information, which is also scalable to more than two views. 
\textcolor{black}{Motivated by the work of \cite{kleinman2022gacs} on decomposing random variables into unique and common components using Gács-Körner common information with variational relaxation, we propose a novel multi-view learning framework. This framework aims to mitigate issues related to the uncertainty introduced by variational approximations. } 

To this end, we initially formulate a mathematically rigorous definition of deterministic common information for multi-view data, by drawing upon the G\'acs-K\"orner common information definition~\textcolor{black}{\cite{kleinman2022gacs, maurer1993secret, gacs1973common}} in information theory literature~\cite{cover1999elements}.    
This defined common information is operationalized through neural networks, culminating in a novel deterministic common and unique multi-view information learning framework.
Leveraging this definition, our framework is capable of extracting common information from multi-view data. 
The employment of total correlation~\cite{watanabe1960information, yu2019multivariate} ensures the independence of common and unique information across different views. 
Effectively, the framework segregates common and unique information, thereby harmonizing complementarity and consensus among views.
The contributions of this work include:
\begin{itemize}
    \item We formulate a mathematically rigorous definition of common information for multi-view data, rooted in the principles of information theory.
    \item \textcolor{black}{We present a novel multi-view learning framework that utilizes our definition of common information and the matrix-based R{\'e}nyi's $\alpha$-order total correlation to achieve a representation skilled at distinguishing between common and unique information.}
    \item \textcolor{black}{Our framework is scalable to handle multi-view data involving more than two views.}
    \item Experimental results substantiate the efficacy of the proposed framework.
\end{itemize}

\section{Related Work}
\subsection{Multi-view Learning: From Classical to Information Theoretical Approaches}   
An extensive body of literature exists on multi-view learning. 
For readers interested in classical algorithms, we recommend referring to these surveys~\cite{sun2013survey, xu2013survey, zhao2017multi}. 
For deep learning-based multi-view learning techniques, the studies in~\cite{yan2021deep, qi2021review} provide valuable insights. 
This section primarily focuses on methods pertinent to our research, highlighting both traditional multi-view learning approaches and those rooted in information theoretic principles.

One traditional approach primarily employs least squares regression to construct a transformation matrix, denoted as $\mathcal{W}=\{{\mathcal{W}}_i\}_{i=1}^v$, which is utilized for the classification of the observations $\{\mathcal{X}^{(i)}\}^v_{i=1}$ from a total of $v$ views. 
The optimization objective of this process is as follows:
\begin{equation}\label{eq:classical}
    \textcolor{black}{\arg \min_{\mathcal{W},\mathcal{B}}\sum_{i=1}^{v}\left \| \mathcal{W}^{T}_i \mathcal{X}^{(i)}+\mathcal{B}-\mathcal{Y} \right \|_F^2 + \lambda \sum_{i=1}^{v} R (\mathcal{W}_i),}
\end{equation}
where $\mathcal{Y}$ is the corresponding label information of observations, $\mathcal{B}$ is an intercept vector, $\lambda$ is a trade-off parameter, $R$ is a constraint on the transformation matrix, \textcolor{black}{and $\|\cdot\|_F$ denotes the Frobenius norm of a matrix}.
\textcolor{black}{The first term in Equation~\eqref{eq:classical} represents a discriminative regression target, while the second term serves as a constraint on the transformation matrix.}
Various strategies employ different constraints $R$ to fulfill manifold objectives.
In reference~\cite{yang2019adaptive}, a discriminative regression target is utilized, and the adaptive weight parameter is integrated into the transformation matrix to achieve adaptive-weighting discriminative regression for multi-view classification. 
Reference~\cite{wang2013multi} constrains the transformation matrix using the $\ell_{2,1}$-norm and $\ell_1$-norm to achieve structural sparsity and integrate all features.

In addition, the use of an information-theoretical framework in multi-view learning has become increasingly prevalent~\cite{xu2014large, zhang2022multi, federici2020learning}, especially with the significant rise of the information bottleneck principle, achieving a harmonious balance between model intricacy and precision. 
Specifically, the information bottleneck principle guides the model to extract concise and accurate representations $\mathcal{Z}$ from the data $\{\mathcal{X}^{(i)}\}^v_{i=1}$ for each view.
In a supervised setting, this principle plays a pivotal role in equilibrating the model's complexity and accuracy by managing the tradeoff between sufficiency (i.e., the model's performance on the task, as quantified by mutual information $I(\mathcal{Y}, \mathcal{Z})$) and minimality (i.e., the complexity of the representation, as assessed by mutual information $I(\{\mathcal{X}^{(i)}\}^v_{i=1}, \mathcal{Z})$)~\cite{gilad2003information}.
The expression is as follows: 
\begin{equation}
    \arg \max_{\mathcal{Z}} {I(\mathcal{Y}, \mathcal{Z}) - \beta I(\{\mathcal{X}^{(i)}\}^v_{i=1}, \mathcal{Z})}.
\end{equation}
\textcolor{black}{In a two-view unsupervised scenario involving} $\mathcal{X}_1$ and $\mathcal{X}_2$, the information bottleneck approach can be expressed as~\cite{federici2020learning}:
\begin{equation}\label{eq:federici}
    \arg \max_{\theta, \psi; \beta} I(\mathcal{Z}_1;\mathcal{Z}_2) - \beta D_{SKL}(p_{\theta}(\mathcal{Z}_1 | \mathcal{X}_1) || p_{\psi}(\mathcal{Z}_2 | \mathcal{X}_2)), 
\end{equation}
where $I(\mathcal{Z}_1;\mathcal{Z}_2)$ is mutual information between two representations, and $D_{SKL}$ is the symmetrized KL divergence, $\theta, \psi$ are encoders, while the coefficient $\beta$ defines the trade-off between sufficiency and robustness of the representation.

However, current approaches lack a clear and rigorous mathematical foundation for identifying common information across multiple views. 
In addition, counter-intuitive higher order interaction between multiple views makes the unsupervised method non-trivial to generalize to more than two views~\cite{federici2020learning}.
In contrast, our proposed method establishes a well-defined mathematical approach for common information in multi-view. 

\subsection{G\'acs-K\"orner Common Information}
Prior to mathematically introducing G\'acs-K\"orner common information, we present an illustrative example to facilitate an intuitive understanding.
Consider two statistical descriptions, $\mathcal{S}_1$ and $\mathcal{S}_2$, each characterizing dissimilar sets of images.
$\mathcal{S}_1$ pertains to images featuring airplanes and blue skies, while $\mathcal{S}_2$ pertains to images featuring unicorns and blue skies. 
The common information between $\mathcal{S}_1$ and $\mathcal{S}_2$ intuitively corresponds to the number of bits required to describe this shared feature, namely the blue sky~\cite{yu2022common}.
Yet, a fundamental question arises: how can we quantitatively and precisely define the intrinsic similarity or common information between two correlated random variables like  $\mathcal{S}_1$ and $\mathcal{S}_2$?

The G\'acs-K\"orner common information, also known as ``zero error information"~\cite{wolf2004zero}, is defined as the random variable $\mathcal{T}$ determined by two input random variables $\mathcal{S}_1$ and $\mathcal{S}_2$ via deterministic functions but carries the maximal entropy~\cite{gacs1973common}:
\begin{equation}\label{eq:original_GK}
\begin{split}
      & GK(\mathcal{S}_1;\mathcal{S}_2) :=  \max_{\mathcal{T}} H(\mathcal{T}) \\ &  s.t. \quad \mathcal{T}=\varphi_1(\mathcal{S}_1)=\varphi_2(\mathcal{S}_2)
\end{split}
\end{equation}
Here, $\varphi_i, (i=1,2)$, signifies a deterministic function for the input random variable, and $H(\mathcal{T})$ denotes the entropy of the random variable $\mathcal{T}$.
Serving as a lower bound to the mutual information, the G\'acs-K\"orner common information also addresses the notion of shared information between sources, an aspect not fully captured by mutual information.
The aim of this definition is to create a formal and operational framework for quantifying commonality between sources~\cite{kleinman2022gacs}.

The quantification of common information has crucial implications for information coding, computer science, and cryptography~\cite{maurer1993secret, yu2022common}. 
Although it has recently garnered attention in the field of machine learning, no methods currently exist to directly compute the G\'acs-K\"orner common information from high-dimensional samples in a deterministic manner.
A variational relaxation G\'acs-K\"orner approach has been proposed for extracting common information in two-view settings~\cite{kleinman2022gacs}. 
\textcolor{black}{However, this variational common information (VCI) approach exhibits certain limitations, and our method distinguishes itself from it.
First, our method utilizes a deterministic model, eliminating the need for variational approximation and thereby reducing uncertainty induced by such approximations.
Second, our technique enforces constraints on the independence of common and unique information, ensuring strict adherence to theoretical principles.
This is an aspect absent in the VCI approach, which consequently renders it less rigorous.
In contrast, our approach is universally applicable to multi-view supervised scenarios, irrespective of the number of views.}

\subsection{Matrix-based R{\'e}nyi's $\alpha$-order Entropy}\label{sec:Matrix-based Entropy}
\textcolor{black}{This section presents the methodology for estimating the entropy of a variable $\mathbf{x}$, denoted as $H(\mathbf{x})$, and the total correlation among variables $(\mathbf{x}_1,\dots,\mathbf{x}_{v})$, denoted as $\text{TC}(\mathbf{x}_1,\dots,\mathbf{x}_{v})$. This is achieved through the use of the matrix-based R{\'e}nyi's $\alpha$-order entropy functional.
}

\textcolor{black}{Following~\cite{yu2019multivariate, giraldo2014measures}, the entropy of variable $\mathbf{x}$ can be defined over the eigenspectrum of a (normalized) Gram matrix $K_{{\mathbf{x}}}\in \mathbb{R}^{N\times N}$ ($K_{{\mathbf{x}}}(m,n)=\kappa({\mathbf{x}}^{m}, {\mathbf{x}}^{n})$, where $\kappa$ is a Gaussian kernel) as:
\begin{equation}\label{eq:Renyi_entropy}
\begin{split}
    &H_{\alpha}(\mathcal{A}_{\mathbf{x}})=\frac{1}{1-\alpha}\log_2 \left(\operatorname{tr} (\mathcal{A}_{\mathbf{x}}^{\alpha})\right) \\
    &=\frac{1}{1-\alpha}\log_{2}\left(\sum_{m=1}^{N}\lambda _{m}(\mathcal{A}_{\mathbf{x}})^{\alpha}\right),  
\end{split}
\end{equation}
where $\alpha\in (0,1)\cup(1,\infty)$. 
$\mathcal{A}_{\mathbf{x}}$ is the normalized version of $K_{\mathbf{x}}$, i.e., $\mathcal{A}_{\mathbf{x}}=K_{\mathbf{x}}/\operatorname{tr}(K_{\mathbf{x}})$. $\lambda_{m}(\mathcal{A}_{\mathbf{x}})$ denotes the $m$-th eigenvalue of $\mathcal{A}_{\mathbf{x}}$.}

\textcolor{black}{The joint entropy for $\{\mathbf{x}_1, \mathbf{x}_2, \dots, \mathbf{x}_v\}$ can be defined as:
\begin{equation}\label{eq:joint_entropy}
    H_\alpha (\mathcal{A}_{\mathbf{x}_1},\mathcal{A}_{\mathbf{x}_2},\dots, \mathcal{A}_{\mathbf{x}_v}) = H_{\alpha}\left(\frac{\mathcal{A}_{\mathbf{x}_1} \circ \mathcal{A}_{\mathbf{x}_2} \circ \dots  \circ \mathcal{A}_{\mathbf{x}_v}}{\operatorname{tr} (\mathcal{A}_{\mathbf{x}_1} \circ \mathcal{A}_{\mathbf{x}_2} \circ \dots  \circ \mathcal{A}_{\mathbf{x}_v})}\right),
\end{equation}
where $\circ$ denotes the Hadamard (or element-wise) product.}

\textcolor{black}{The matrix-based R{\'e}nyi's $\alpha$-order total correlation $\text{TC}(\mathbf{x}_1, \mathbf{x}_2, \dots, \mathbf{x}_v)$ is defined by extending the concept that mutual information is the KL divergence between the joint distribution and the product of marginals. 
It quantifies the total amount of dependence among the variables. 
Formally, $\text{TC}(\mathbf{x}_1, \mathbf{x}_2, \dots, \mathbf{x}_v)$ can be expressed in terms of individual entropies and joint entropy as:
\begin{equation}\label{eq:Renyi_TC}
\begin{split}
    & \text{TC}_{\alpha}(\mathbf{x}_1, \mathbf{x}_2, \dots, \mathbf{x}_v)=H_{\alpha}(\mathcal{A}_{\mathbf{x}_1}) + H_{\alpha}(\mathcal{A}_{\mathbf{x}_2}) + \dots \\
    & + H_{\alpha}(\mathcal{A}_{\mathbf{x}_v})  - H_{\alpha}\left(\frac{\mathcal{A}_{\mathbf{x}_1} \circ \mathcal{A}_{\mathbf{x}_2} \circ \dots  \circ \mathcal{A}_{\mathbf{x}_v}}{\operatorname{tr} (\mathcal{A}_{\mathbf{x}_1} \circ \mathcal{A}_{\mathbf{x}_2} \circ \dots  \circ \mathcal{A}_{\mathbf{x}_v})}\right).
\end{split}
\end{equation}}

\textcolor{black}{The differentiability of R{\'e}nyi's $\alpha$-order entropy functional based on matrices has been demonstrated in this work~\cite{yu2021measuring}. 
In practice, common deep learning APIs such as TensorFlow and PyTorch include automatic eigenvalue decomposition.}

\section{Proposed Method}
\subsection{Problem Formulation}
Consider a dataset of $n$ observations across $v$ views, denoted as $\{\mathcal{X}^{(i)}\}^v_{i=1}$, where $\mathcal{X}^{(i)}\in \mathbb{R}^{n \times d^{(i)}}$ represents the $i$-th view of data and $d^{(i)}$ signifies its feature dimension. 
Let $\mathcal{C}\in \mathbb{R}^{n \times d_{\mathcal{C}}}$ denote the common features across all views, with $d_{c}$ representing the dimension of the common feature.
The unique features are denoted as $\mathcal{U}=\{\mathcal{U}^{(1)},\dots,\mathcal{U}^{(v)}\}$, where $\mathcal{U}^{(i)}\in \mathbb{R}^{n \times d_{\cal U}^{(i)}}$ represents the unique feature from the individual $i$-th view and $d_{\tiny{\cal U}}^{(i)}$ is the corresponding feature dimension. 
The objective is to extract a set of common features $\mathcal{C}$ with maximal entropy and a set of unique features $\mathcal{U}$ from the variable collection $\{\mathcal{X}^{(i)}\}^v_{i=1}$ while ensuring maximum independence between them.

\subsection{Common Information for Multi-view Data}
Multi-view data are typically characterized by both common and unique information.
In this study, we present a definition to quantify the common information within multi-view data.
Importantly, to accommodate data with more than two views, our definition extends the concept of G\'acs-K\"orner common information, as illustrated in Equation~(\ref{eq:original_GK}). 
We denote this generalized concept as multi-view G\'acs-K\"orner common information.
\begin{definition}\label{def:ci}
    Consider a multi-view data $\{\mathcal{X}^{(i)}\}^v_{i=1}$ with $\mathcal{X}^{(i)}$ denoting data from the $i$-th view, an entropy measure $H$,  and $v$ deterministic functions $\{\phi^{(i)}\}_{i=1}^{v}$ with $\phi^{(i)}: {\mathbb{R}}^{d^{(i)}} \to {\mathbb{R}}^{d_{\mathcal{C}}} $, the multi-view G\'acs-K\"orner common information is:
    \begin{equation}\label{eq:extend_GK} 
        \begin{split}
            & GK(\mathcal{X}^{(1)};\dots;\mathcal{X}^{(v)}) :=  \max_{\mathcal{C}} H(\mathcal{C}) \\ &  s.t. \quad \mathcal{C}=\phi^{(i)}(\mathcal{X}^{(i)}),\quad i=1,\dots,v.
        \end{split}
    \end{equation} 
\end{definition}
This definition delineates the procedure for extracting common information from multi-view data. 
For multi-view data $\{\mathcal{X}^{(i)}\}^v_{i=1}$, a common variable $\mathcal{C}$ can be derived from one view's data, $\mathcal{X}^{(1)}$, using a deterministic function $\phi^{(1)}$.
This common variable $\mathcal{C}$ should be in congruence with the variable derived from any other view $\mathcal{X}^{(i)}$ using a corresponding deterministic function $\phi^{(i)}$.
When the common information encompasses the maximum amount of information, i.e., $\max H(\mathcal{C})$, it signifies that it has captured the precise amount of common information among the multi-view data. 
That is why we maximize the entropy of the common features $\mathcal{C}$.

Regarding the implementation of neural network algorithms, $\phi^{(i)}$  serves as an encoder for the $i$-th view.
$\mathcal{C}$ represents a common feature extracted from different views by their respective encoders, aiming to maximize the information embedded in this common feature, that is, $\max H(\mathcal{C})$.

\subsection{Common and Unique Multi-view Information Learning Framework}\label{sec:CUMI framework}
\begin{figure*}[t!]
    \centering
    \includegraphics[width=0.8\textwidth]{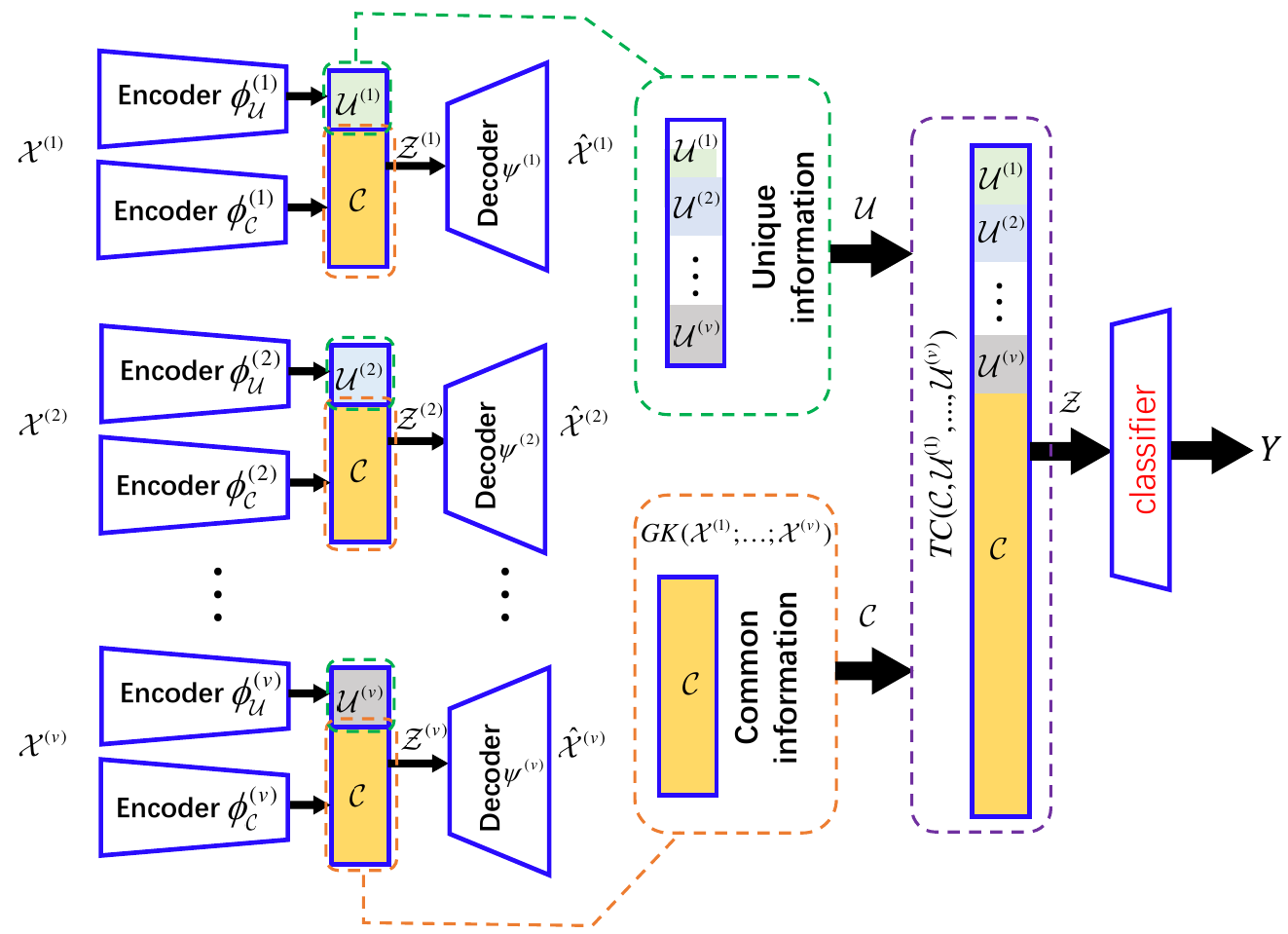}
    \caption{Schematic representation of the common and unique multi-view information (CUMI) learning Framework. 
    The CUMI framework is designed to learn a joint representation, denoted as $\mathcal{Z}$, which comprises common features $\mathcal{C}$ and unique features $\mathcal{U}$.
    The encoder $\phi_{\mathcal{C}}$ is responsible for extracting common features, guided by our definition of the multi-view common information criterion as per Equation (\ref{eq:extend_GK}).
    Unique features pertaining to the $i$-th view, represented as $\mathcal{U}^{(i)}$, are extracted by the independent encoder $\phi_{\mathcal{U}}^{{(i)}}$, which is coupled with a reconstruction network $\psi_{(i)}$.
    To maintain the independence of the components in $\mathcal{Z}$, we introduce a total correlation constraint, denoted as $\text{TC}$, to ensure the independence of each term.}
    \label{fig:framework}
\end{figure*}

Informed by our definition of common information in multi-view data, we put forth a \textbf{C}ommon and \textbf{U}nique \textbf{M}ulti-view \textbf{I}nformation learning framework (\textbf{CUMI}), depicted in Figure~\ref{fig:framework}.
The CUMI framework comprises two main components: the extraction of common and unique information.
The common information $\mathcal{C}$, which encapsulates common features across various views, is extracted in accordance with our definition of common information. 
The extraction of unique information $\mathcal{U}$, or distinctive features, is governed by the constraints of the reconstruction network.

To extract common features $\mathcal{C}$ from multi-view data, we employ a common encoder $\phi_\mathcal{C}^{(i)}$ for each view.
We follow our definition of common information as per Equation (\ref{eq:extend_GK}), and maximize the entropy $H(\mathcal{C})$, ensuring that the common features obtained from different views by the common encoder are consensus and sufficient.
The common information $\mathcal{C}$ across multiple views is obtained by randomly selecting $\mathcal{C}$ from a given view, following a uniform distribution. This implies that each view has an equal probability of being chosen. 
The backpropagation of the loss function ensures the convergence of the common information across all views, leading to consensus.

In addition to capturing common information, we also consider the compensatory effect of unique information.
We use individual encoders $\phi_\mathcal{U}^{(i)}$ to extract unique features, which are then combined with common features to form the reconstruction network $\psi$.
{\color{black}
To extract the unique information $\mathcal{U}^{(i)}$ from each view, we assume that the combination of $\mathcal{C}$ and $\mathcal{U}^{(i)}$ guarantees faithfully reconstruction of view-specific data.  
We employ the mean squared error ({MSE}) as the loss for reconstruction:
\begin{equation}\label{eq:loss_recon}
         \argmin_{\phi, \psi}  \sum_{i=1}^{v} \text{MSE} (\mathcal{X}^{(i)};\hat{\mathcal{X}}^{(i)} ),
\end{equation}
where $\phi, \psi$ are parameters of each view's encoder and decoder. 
By utilizing the specialized reconstruction network, the extracted unique and common information is used to reconstruct the original data, thereby ensuring the framework’s practical validity from an engineering perspective.

Further, we enforce the mutual independence between the extracted common information $\mathcal{C}$ and the unique information $\mathcal{U}^{(i)}$ from each view by minimizing a total correlation (TC)\footnote{The total correlation (TC) is a multivariate generalization of mutual information, which quantifies the total dependence between multiple variables $(\mathcal{C},\mathcal{U}^{(1)},\dots,\mathcal{U}^{(v)})$ and can be expressed as:
$\text{TC}(\mathcal{C},\mathcal{U}^{(1)},\dots,\mathcal{U}^{(v)}) = H(\mathcal{C}) +\sum_{i=1}^v H(\mathcal{U}^{(i)}) - H(\mathcal{C},\mathcal{U}^{(1)},\dots,\mathcal{U}^{(v)})$,
where $H$ refers to entropy or joint entropy. 
TC reduces to zero if and only if ($\mathcal{C},\mathcal{U}^{(1)},\dots,\mathcal{U}^{(v)}$) are independently to each other~\cite{yu2021measuring}.} term~\cite{ watanabe1960information} $\text{TC}(\mathcal{C},\mathcal{U}^{(1)},\dots,\mathcal{U}^{(v)})$.
The total correlation constraint theoretically guarantees the efficacy of our framework, due to Proposition~\ref{prop:independent}.

\begin{globalthm}\label{prop:independent}
    Define 
    \begin{equation}
        \mathcal{Z}_1 = (\mathcal{C}, \mathcal{U}^{(1)}), \, \, \dots, \, \, \mathcal{Z}_v = (\mathcal{C}, \mathcal{U}^{(v)})
    \end{equation}
    where $\mathcal{C}$, $\mathcal{U}^{(1)}$,$\dots$, $\mathcal{U}^{(v)}$ are mutually independent. 
    \textcolor{black}{Then, for a set of any invertible transformations $\{f_i\}_{i=1}^v$, the random variable $\mathcal{Z}^*$ optimized from Equation \eqref{eq:prop1} represents the common information $\mathcal{C}$ that is concealed within all views.}
    \begin{equation}\label{eq:prop1}
        \arg \max_{\mathcal{Z}} GK(f_{1}(\mathcal{Z}^{(1)}), \dots, f_{v}(\mathcal{Z}^{(v)}))
    \end{equation}
\end{globalthm}

If $v=2$, the proposition has already been proven in \textcolor{black}{\cite{wolf2004zero, kleinman2022gacs}}. 
\textcolor{black}{We extend this result straightforwardly to cases with more than two views and furnish the proof following}.  

\textit{Proof.} \textcolor{black}{Note that if $f$ is the identity transformation $f(\mathcal{Z})=\mathcal{Z}$, then $\arg \max_{\mathcal{Z}} GK(\mathcal{Z}^{(1)}, \dots, \mathcal{Z}^{(v)})$ is $\mathcal{C}$.}
If $f$ is an invertible transformations, suppose that $g_{i} (i=1,\dots,v)$ are the functions satisfying $\mathcal{Z}=g_{1}(\mathcal{Z}^{(1)})=\dots=g_{v}(\mathcal{Z}^{(v)})$ corresponding to $GK(\mathcal{Z}^{(1)}, \dots, \mathcal{Z}^{(v)})$.
Then the functions corresponding to $GK(f_{1}(\mathcal{Z}^{(1)}$, $\dots$, $f_{v}(\mathcal{Z}^{(v)})$ will be $\mathcal{Z}=g_1 \circ f_1^{-1}(f_1(\mathcal{Z}^{(1)}))=\dots= g_v \circ f_v^{-1}(f_v(\mathcal{Z}^{(v)}))$ and the random variable $\mathcal{Z}$ is equivalent.

This concludes the proof\textcolor{black}{\footnote{\textcolor{black}{As a straightforward extension of the two-view case presented in \cite{kleinman2022gacs}, Proposition \ref{prop:independent} demonstrates that the results hold for three or more views. 
}}}.

The proposition demonstrates that if a set of common variables ($\mathcal{C}$) is consensual across views and independent of unique variables ($\mathcal{U}^{(1)},\dots,\mathcal{U}^{(v)}$), then the G\'acs-K\"orner common random variable corresponds to the common information for multi-view data. 
The proposition emphasizes the importance of imposing the total correlation constraint, $\text{TC}(\mathcal{C},\mathcal{U}^{(1)},\dots,\mathcal{U}^{(v)})$. 
This constraint, subsequently, ensures the theoretical rigor of our framework.

Therefore, the final and overall objective of our framework can be expressed as:
\begin{equation}\label{eq:final_objective}
\begin{split}
    &\argmin_{\phi, \psi} \texttt{CE}(\mathcal{Y};\mathcal{\hat{Y}}) + \sum_{i=1}^{v} \text{MSE}(\mathcal{X};\mathcal{\hat{X}}) - \beta H(\mathcal{C}) \\ 
    & + \gamma \text{TC}(\mathcal{C},\mathcal{U}^{(1)}, \dots, \mathcal{U}^{(v)})
\end{split}
\end{equation}
where $\beta$ and $\gamma$ are regularization parameters for the common information and total correlation terms, respectively. 
\textcolor{black}{Additionally, $\texttt{CE}$ denotes cross-entropy utilized in classification tasks.}
The main challenge in optimizing Equation~(\ref{eq:final_objective}) lies in the fact that the exact computation of $H(\mathcal{C})$ and $\text{TC}(\mathcal{C},\mathcal{U}^{(1)},\dots,\mathcal{U}^{(v)})$ is almost impossible or intractable due to the high dimensionality of the data.
}

\textcolor{black}{In this study, we tackle the issue of estimating $H(\mathcal{C})$ and total correlation $\text{TC}(\mathcal{C},\mathcal{U}^{(1)},\dots,\mathcal{U}^{(v)})$ by employing the matrix-based R{\'e}nyi's $\alpha$-order entropy functional.
Specifically, given a minibatch of $N$ samples, we have a set of representations $\{\mathbf{c}_m, \mathbf{u}^1_m, \dots, \mathbf{u}^v_m\}_{m=1}^{N}$. 
We can regard both $\mathbf{c}$ and $\mathbf{u}$ as random vectors.
Building upon Section~\ref{sec:Matrix-based Entropy}, the entropy of the variable $\mathbf{c}$ is defined as follows:
\begin{equation}\label{eq:Renyi_entropy_our}
\begin{split}
    &H_{\alpha}(\mathcal{A}_{\mathbf{c}})=\frac{1}{1-\alpha}\log_{2}\left(\sum_{m=1}^{N}\lambda _{m}(\mathcal{A}_{\mathbf{c}})^{\alpha}\right),  
\end{split}
\end{equation}
where $\alpha\in (0,1)\cup(1,\infty)$. }

\textcolor{black}{Formally, $\text{TC}(\mathbf{c}, \mathbf{u}^1, \dots, \mathbf{u}^v)$ can be expressed as:
\begin{equation}\label{eq:Renyi_TC_our}
\begin{split}
    & \text{TC}_{\alpha}(\mathbf{c}, \mathbf{u}^1, \dots, \mathbf{u}^v)=H_{\alpha}(\mathcal{A}_{\mathbf{c}}) + H_{\alpha}(\mathcal{A}_{\mathbf{u}^1}) + \dots \\
    & + H_{\alpha}(\mathcal{A}_{\mathbf{u}^v})  - H_{\alpha}\left(\frac{\mathcal{A}_{\mathbf{c}} \circ \mathcal{A}_{\mathbf{u}^1} \circ \dots  \circ \mathcal{A}_{\mathbf{u}^v}}{\operatorname{tr} (\mathcal{A}_{\mathbf{c}} \circ \mathcal{A}_{\mathbf{u}^1} \circ \dots  \circ \mathcal{A}_{\mathbf{u}^v})}\right).
\end{split}
\end{equation}}

\subsection{Verifying the Rationality of Multi-View Common Information}
\begin{globalthm}\label{prop:cumi}
(CUMI discovers the common and unique information.) 
Optimization through Equation~\eqref{eq:final_objective} will discover latents $ {\cal {Z}}$ = $({\cal{C}}_{*}$, ${\cal {U}}_{*}^{(1)}$, ${\cal {U}}_*^{(2)}$,$\cdots$, ${\cal{U}}_*^{(v)})$ where ${\cal{C}_*}$ is the common random variable that maximizes the multi-view G\'acs-K\"orner common information in Equation~\eqref{eq:extend_GK}, and ${\cal { U}_*}^{(i)}$ is the unique information of the $i$-th view, which maximizes $I({\cal{X}}^{(i)}; {\cal {C}}, {\cal {U}}^{(i)})$.
\end{globalthm}
\textit{Proof.} 
In accordance with Equation~\eqref{eq:final_objective}, optimization through the CUMI framework will culminate in
\begin{equation} \label{eq:proof_common}
    \max H(\mathcal{C}) \quad  s.t. \quad \mathcal{C}=\phi^{(i)}(\mathcal{X}^{(i)}),\quad i=1,\dots,v.
\end{equation}
This is just correspondent to the claim of the G\'acs-K\"orner common information in Equation~\eqref{eq:extend_GK}.

The optimization of Equation~\eqref{eq:final_objective} will satisfy Equation~\eqref{eq:loss_recon} which implies
\begin{equation}\label{eq:entropy_recon_1}
    H({\cal X}^{(i)})=H(\hat{\cal X}_*^{(i)}). 
\end{equation}
On the other hand, modeling our framework will induce 
\begin{equation}
    \begin{array}{l}
    {\cal \hat{X}}^{(i)}=\psi ^{(i)}({\mathcal{C}}, {\cal U}^{(i)})=\psi ^{(i)}(\phi_{\mathcal{C}} ^{(i)} ({{\cal{X}}^{(i)}}), \phi_{\cal U} ^{(i)} ({{\cal{X}}^{(i)}})),  
    \end{array}
\end{equation}
from which we can derive
\begin{equation}\label{eq:entropy_recon_2}
    H({\cal{X}}^{(i)}) \ge H({\mathcal{C}}, {\cal U}^{(i)}) \ge H({\cal \hat{X}}^{(i)})
\end{equation}
Now, through \textcolor{black}{Equations}~\eqref{eq:entropy_recon_1} and \eqref{eq:entropy_recon_2} we can determine that
\begin{equation}
   I({{{\cal X}}^{(i)}};{{\mathcal{C}}_*},{{{\cal U}}_*^{(i)}})=H({{\mathcal{C}}_*},{{{\cal U}}_*^{(i)}})=H({\cal X}^{(i)})
\end{equation}
and ${{\cal U}}_*^{\left( i \right)}$ satisfies ${{\cal U}}_*^{\left( i \right)} = \mathop {\max }\limits_{{{{\cal U}}^{(i)}}} I({{{\cal X}}^{(i)}};{{\mathcal{C}}},{{{\cal U}}^{(i)}})$.

This concludes the proof.

\subsection{Discussion on Key Points}
This section provides a discussion on the key points of our proposed approach.

\subsubsection{Common Information Captures More Than Mutual Information}
This section elucidates the distinction between mutual information and common information, and provides justification for employing multi-view G\'acs-K\"orner common information.

\textcolor{black}{We reiterate the example provided by \cite{kleinman2022gacs}}: $ \mathcal{Z}_1 = (\mathcal{C}, \mathcal{U}^{(1)}), \mathcal{Z}_2 = (\mathcal{C}, \mathcal{U}^{(2)})$, where $\mathcal{C}, \mathcal{U}^{(i)}$ are independent. 
In this context, $\mathcal{C}$ (for instance, is a binary variable 0 or 1) denotes the common information shared between the views, while $\mathcal{U}^{(1)}$ and $\mathcal{U}^{(2)}$ (for instance, are small amounts of correlated Gaussian noise) represent the unique information contained in each view. 
It is assumed that $\mathcal{U}^{(1)}$ and $\mathcal{U}^{(2)}$ are correlated but unique\footnote{https://openreview.net/forum?id=e4XidX6AHd\&noteId=HqfjeyTDaR.}. 

The mutual information between $\mathcal{Z}_1$ and $\mathcal{Z}_2$ can be expressed as:

\begin{equation}\label{eq:CInotMI}
    I(\mathcal{Z}_1;\mathcal{Z}_2) = H(\mathcal{C}) + I(\mathcal{U}^{(1)};\mathcal{U}^{(2)})
\end{equation}

\textit{Proof.} Let $\mathcal{Z}_1 = (\mathcal{C}, \mathcal{U}^{(1)})$ and $\mathcal{Z}_2 = (\mathcal{C}, \mathcal{U}^{(2)})$, where $\mathcal{C}$ and $\mathcal{U}^{(i)}$ are independent. The mutual information between $\mathcal{Z}_1$ and $\mathcal{Z}_2$ can be calculated as follows:
\begin{equation}
\begin{split}
    I(\mathcal{Z}_1;\mathcal{Z}_2) = & H(\mathcal{Z}_1) + H(\mathcal{Z}_2) - H(\mathcal{Z}_1, \mathcal{Z}_2) \\
                                   = & H(\mathcal{Z}_1) + H(\mathcal{Z}_2) - H(\mathcal{Z}_1) - H(\mathcal{Z}_2 | \mathcal{Z}_1 ) \\
                                   = & H(\mathcal{Z}_2) - H(\mathcal{Z}_2 | \mathcal{Z}_1 )\\
                                   = & H(\mathcal{C}) + H(\mathcal{U}^{(2)}) - H((\mathcal{C}, \mathcal{U}^{(2)})| (\mathcal{C}, \mathcal{U}^{(1)}) )    \\ 
                                   = & H(\mathcal{C}) + H(\mathcal{U}^{(2)}) - H(\mathcal{U}^{(2)}| \mathcal{U}^{(1)})     \\ 
                                   = & H(\mathcal{C}) + H(\mathcal{U}^{(2)}) - H(\mathcal{U}^{(2)},\mathcal{U}^{(1)}) + H(\mathcal{U}^{(1)})\\
                                   = & H(\mathcal{C}) + I(\mathcal{U}^{(1)};\mathcal{U}^{(2)})
\end{split}
\end{equation}
Here, $H$ denotes entropy or joint entropy. 

Under these circumstances, the multi-view common information defined by G\'acs-K\"orner can be identified as $H(\mathcal{C})$ (for instance, capture the ture $\mathcal{C}$ by threshold). 
Conversely, from Equation (\ref{eq:CInotMI}), it is evident that if the correlation between $\mathcal{U}^{(1)}$ and $\mathcal{U}^{(2)}$ is sufficiently strong, the value of $I(\mathcal{Z}_1;\mathcal{Z}_2)$ will be significantly greater than $H(\mathcal{C})$. 
Therefore, our method identifies common component rather than merely assessing correlations, which can be influenced by spurious factors.
\subsubsection{Unique Information is Decoupled from Common Information}
To decouple $\mathcal{U}^{(i)}$ from $\mathcal{C}$, we minimize the total correlation among ($\mathcal{C}$, $\mathcal{U}^{(1)}$, ..., $\mathcal{U}^{(v)}$).
Considering two views $\mathcal{X}_1$ and $\mathcal{X}_2$, our framework reconstructs $\mathcal{X}_1$ using ($\mathcal{U}^{(1)}$, $\mathcal{C}$) and $\mathcal{X}_2$ using ($\mathcal{U}^{(2)}$, $\mathcal{C}$).
The fact that $\mathcal{U}^{(1)}$ does not contribute to the reconstruction of $\mathcal{X}_2$ and vice versa suggests that $\mathcal{U}$ contains unique and does not include common information.

\section{Experiments}
In our experiments, we juxtapose the proposed CUMI method against existing state-of-the-art multi-view representation learning algorithms. 
These comparisons are conducted on both synthetic data and several real-world multi-view datasets.

\subsection{Validation on Synthetic Data}
This section employs a two-view synthetic dataset, provided by \cite{salzmann2010factorized}, to validate the efficacy of the proposed methodology. 
The methodology is designed to successfully separate the common and unique information.

We examine the competitive performance of CCA \cite{chaudhuri2009multi} and DCCA \cite{andrew2013deep}. 
In addition, we also test the performance of the unsupervised multi-view approach MIB~\cite{federici2020learning},  and VCI~\cite{kleinman2022gacs}.
\textcolor{black}{However, these two approaches are only designed for two views, in which the way of scaling to more than two views are still unknown. 
Hence, we ignore our comparison to MIB and VCI in real-world data, which usually contains more than 4 views. }

\subsubsection{Generation of Synthetic Data}
We generate 100 samples from two data streams that include both common and unique information. 
The ground-truth common representation is denoted by $\mathbf{c}=\sin(2 \pi t)$. 
On the other hand, $\mathbf{u}^{(1)}=\cos(\pi ^2 t)$ and  $\mathbf{u}^{(2)}=\cos(\sqrt{5} \pi t)$ serve as distinct ground-truth unique representations, with $t$ being uniformly distributed in the interval $(-1, 1)$. 
These unique representations exhibit different frequencies.
Consequently, the observational data are generated by randomly projecting the common and unique representations into a 20-dimensional space, denoted as $\mathcal{X}^{(1)}$=$f[\mathbf{c},\mathbf{u}^{(1)}]$+$noise$ and $\mathcal{X}^{(2)}$=$f[\mathbf{c},\mathbf{u}^{(2)}]$+$noise$. 
Here, the function $f$: $R^2 \mapsto R^{20}$ represents a linear mapping, \textcolor{black}{and $noise$ is introduced in the form of $noise=0.02\sin(3.6\pi t)$.}

\subsubsection{Results on Synthetic Data}
\begin{figure*}[t]
    \centering
    \begin{subfigure}[b]{0.25\textwidth}
        \centering
        \includegraphics[width=\textwidth]{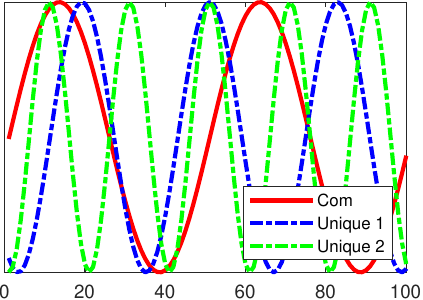}
        \caption{Groundtruth}
        \label{fig:sanity_groundtruth}
    \end{subfigure}
    \hfill
    \begin{subfigure}[b]{0.25\textwidth}
        \centering
        \includegraphics[width=\textwidth]{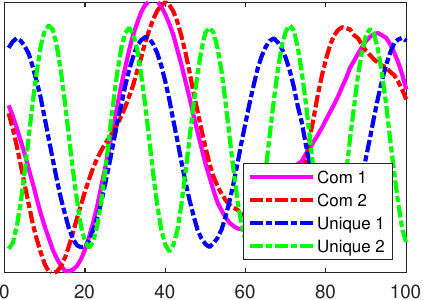}
        \caption{CCA}
        \label{fig:sanity_cca}
    \end{subfigure}
    \hfill
    \begin{subfigure}[b]{0.25\textwidth}
        \centering
        \includegraphics[width=\textwidth]{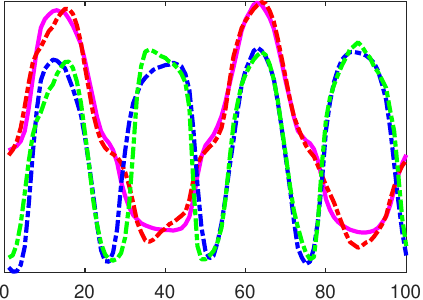}
        \caption{DCCA}
        \label{fig:sanity_dcca}
    \end{subfigure}
    \\
    \begin{subfigure}[b]{0.25\textwidth}
        \centering
        \includegraphics[width=\textwidth]{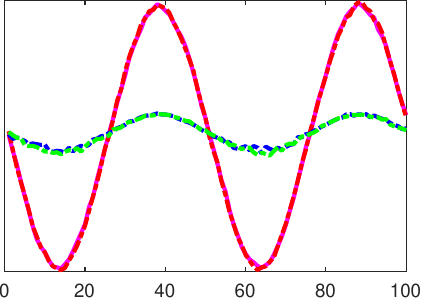}
        \caption{MIB}
        \label{fig:sanity_mib}
    \end{subfigure}
    \hfill 
    \begin{subfigure}[b]{0.25\textwidth}
        \centering
        \includegraphics[width=\textwidth]{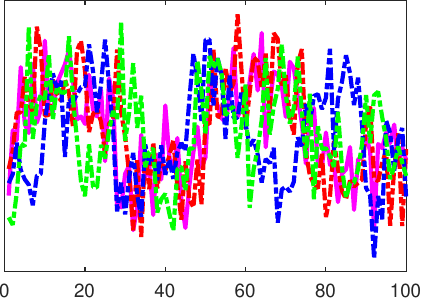}
        \caption{VCI Method}
        \label{fig:sanity_Kleinman}
    \end{subfigure}
    \hfill 
    \begin{subfigure}[b]{0.25\textwidth}
        \centering
        \includegraphics[width=\textwidth]{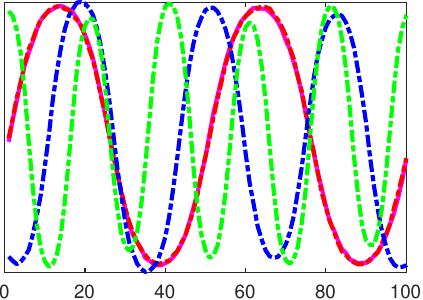}
        \caption{Our Method}
        \label{fig:sanity_our}
    \end{subfigure}
    \caption{Comparison of various methods in discovering common and unique information from multi-view data.
    Ground-truth (\ref{fig:sanity_groundtruth}) displays common (red) and unique information (blue, green). 
    Results for CCA (\ref{fig:sanity_cca}), DCCA (\ref{fig:sanity_dcca}), MIB (\ref{fig:sanity_mib}), VCI (\cite{kleinman2022gacs}), and the proposed method (\ref{fig:sanity_our}) show common (red, magenta) and unique information (blue, green). Our proposed approach closely aligns with the ground truth.}
    \label{fig:sanitycheck}
\end{figure*}

Figure \ref{fig:sanitycheck} illustrates the efficacy of various approaches in discovering common and unique information within multi-view data. 

In Figure \ref{fig:sanity_groundtruth}, the ground-truth underlying representation is depicted, with the red solid line signifying the common information sharing the same frequency. 
The blue and green dashed lines represent the unique information with distinct frequencies for view 1 and view 2, respectively. 
Figures \ref{fig:sanity_cca}, \ref{fig:sanity_dcca}, \ref{fig:sanity_mib}, \ref{fig:sanity_Kleinman} and \ref{fig:sanity_our} present the results of different methods: CCA, DCCA, MIB, VCI, and our proposed approach. 
The red and magenta lines signify the extracted common information from each view, while the blue and green dashed lines denote the recovered unique information. 
The distinction between common and unique information lies in their frequencies. 
It has been observed that CCA fails to accurately capture the frequency of common information, whereas DCCA struggles to precisely represent the frequency of unique information.
Regarding Figure \ref{fig:sanity_mib}, the MIB method exclusively focuses on detecting mutual information between the two views, disregarding the capture of any unique information.
In Figure \ref{fig:sanity_Kleinman}, the inferior performance of the VCI method in recovering the ground truth can be attributed to the absence of independence constraints (TC) on $\mathcal{C}$ and $\mathcal{U}$. 
Additionally, the fluctuation in the curve is evident, which is a result of the inherent uncertainty introduced by the utilization of variational approximation in the VCI method.
As evidenced by the Figure \ref{fig:sanity_our}, our technique effectively retrieves and differentiates common and unique information, achieving a separation effect that closely aligns with the ground truth.
This efficiency primarily results from the proposed definition of multi-view G\'acs-K\"orner common information, which facilitates accurate extraction of common information. 
Moreover, by explicitly minimizing total correlation terms, the common and unique information extracted from each view are compelled to be independent, thus providing a theoretical guarantee for the framework's validity.
By utilizing a specialized reconstruction network, the unique information extracted can be combined with the common features to reconstruct the original data, thereby ensuring the framework's practical validity from an engineering perspective.
Notably, there is no classification loss, $\texttt{CE}$ term, in this experiment. 
This absence demonstrates that the discovery of $\mathcal{C}$ and $\mathcal{U}$ is not driven by the $\texttt{CE}$ term.

\subsubsection{Convergence of $\mathcal{C}$}
In this section, we demonstrate the convergence of $\mathcal{C}$ in our proposed framework. 
We denote $\mathcal{C}^{(i)}$ as the common feature extracted from encoder $\phi_\mathcal{C}^{(i)}$ the $i$-th view. 
We examine the similarity between the feature $\mathcal{C}^{(i)}$ and the common feature $\mathcal{C}$. 
To evaluate the convergence performance, we employ the mean squared error (\textcolor{black}{$\text{MSE}$}) as a measure to quantify the discrepancy between $\mathcal{C}^{(i)}$ and $\mathcal{C}$. 
A decreasing and converging $\text{MSE}$ curve can serve as evidence of the convergence of $\mathcal{C}$, indicating that our method successfully captures the converged common features across different views. 

\begin{figure*}[t]
    \centering
    \begin{subfigure}[b]{0.3\textwidth}
        \centering
        \includegraphics[width=\textwidth]{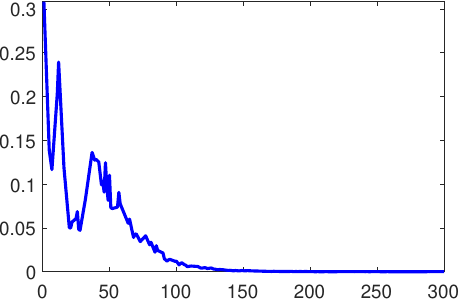}
        \caption{Convergence of $\text{MSE}(\mathcal{C};\mathcal{C}^{(1)})$}
        \label{fig:converge_mse_c1c}
    \end{subfigure}
    \begin{subfigure}[b]{0.3\textwidth}
        \centering
        \includegraphics[width=\textwidth]{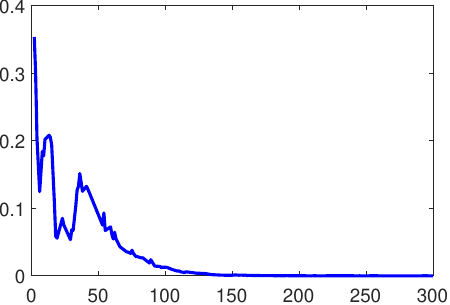}
        \caption{Convergence of $\text{MSE}(\mathcal{C};\mathcal{C}^{(2)})$}
        \label{fig:converge_mse_c2c}
    \end{subfigure}
     \caption{The figure shows the convergence behavior of the $\text{MSE}$ for the common features $\mathcal{C}$. The curves $\text{MSE}(\mathcal{C};\mathcal{C}^{(i)}), i=1,2$ measure the discrepancy between $\mathcal{C}$ and the features $\mathcal{C}^{(i)}$ from different views. 
     The decreasing and converging nature of both curves confirms the successful convergence of $\mathcal{C}$, validating our method's ability to extract consistent common representations from multiple views.
}
    \label{fig:converge_mse}
\end{figure*}

The results in Figure \ref{fig:converge_mse} indicate that both curves exhibit a decreasing trend and eventually converge. This confirms the desired convergence property of our approach. The convergence of the common features provides strong validation for the effectiveness of our proposed multi-view learning algorithm, highlighting its ability to extract common representations from multiple views in a stable and consistent manner.

\subsubsection{Independence between $\mathcal{C}$ and $\mathcal{U}$}
As discussed in Section \ref{sec:CUMI framework}, we have introduced constraints on the independence using total correlation to achieve the decoupling of $\mathcal{C}$ and $\mathcal{U}$. 
In this section, we present the curves depicting the variations of $\text{TC}(\mathcal{C},\mathcal{U}^{(1)}, \dots, \mathcal{U}^{(v)})$, as described in Equation (\ref{eq:Renyi_TC_our}).
Additionally, we assess the independence by utilizing a widely used metric called Hilbert-Schmidt Independence Criterion (HSIC)~\cite{gretton2005measuring,gretton2007kernel}. 
The HSIC quantifies the dependence between two sets of variables by evaluating the similarity of their respective kernel matrices.

\begin{figure*}[t]
    \centering
    \begin{subfigure}[b]{0.3\textwidth}
        \centering
        \includegraphics[width=\textwidth]{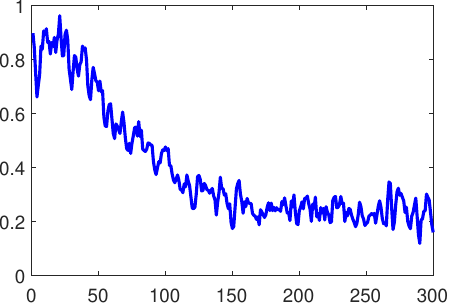}
        \caption{$\text{TC}(\mathcal{C},\mathcal{U}^{(1)}, \mathcal{U}^{(2)})$}
        \label{fig:TC}
    \end{subfigure}
    \begin{subfigure}[b]{0.3\textwidth}
        \centering
        \includegraphics[width=\textwidth]{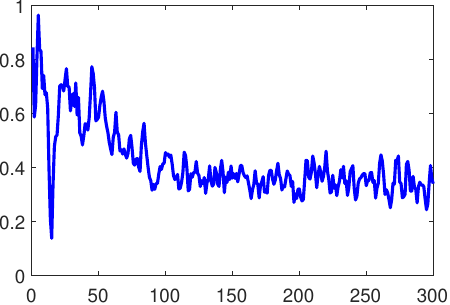}
        \caption{$\text{HSIC}(\mathcal{C},\mathcal{U}^{(1)})$}
        \label{fig:hsic1}
    \end{subfigure}
    \begin{subfigure}[b]{0.3\textwidth}
        \centering
        \includegraphics[width=\textwidth]{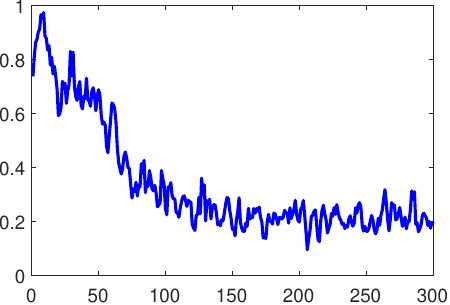}
        \caption{$\text{HSIC}(\mathcal{C},\mathcal{U}^{(2)})$}
        \label{fig:hsic2}
    \end{subfigure}
     \caption{Independence Analysis of $\mathcal{C}$ and $\mathcal{U}$.
     The figure presents the curves of TC and HSIC as measures of the independence between $\mathcal{C}$ and $\mathcal{U}$. 
     These metrics quantify the degree of dependence between $\mathcal{C}$ and each $\mathcal{U}^{(i)}$. 
     The experimental results demonstrate the effectiveness of our constraints in promoting independence between $\mathcal{C}$ and $\mathcal{U}^{(i)}$. 
     The curves exhibit a decreasing trend, providing evidence for the ability of our approach to enforce the desired independence between these variables.
}
    \label{fig:independence}
\end{figure*}

The curves of $\text{TC}(\mathcal{C},\mathcal{U}^{(1)}, \mathcal{U}^{(2)})$, $\text{HSIC}(\mathcal{C},\mathcal{U}^{(1)})$ and $\text{HSIC}(\mathcal{C},\mathcal{U}^{(2)})$ on synthetic datasets are presented in Figure \ref{fig:independence}.
The findings provide evidence of the effectiveness of our constraints, as the curves demonstrate an increasing level of independence between $\mathcal{C}$ and $\mathcal{U}^{(1)}$, $\mathcal{U}^{(2)}$. This validates the capability of our approach in promoting and enforcing independence between these variables.

\subsection{Experiments on Real-world Data}
This section presents the experiments conducted on real-world data to evaluate the performance of our proposed approach.

\subsubsection{Datasets} 
We conduct our experiments on seven datasets.
\textbf{Caltech101-7/20}\footnote{http://www.vision.caltech.edu/Image Datasets/Caltech101/} is a subset of the renowned Caltech101 image classification dataset. 
It consists of 1474/2386 images and includes si x published features: Gabor, wavelet moments (WM), census transform histogram (CENTRIST), histogram of oriented gradients (HOG), greebles image structure texture (GIST), and local binary pattern (LBP)~\cite{fei2004learning}.
\textbf{MSRC}\footnote{http://research.microsoft.com/en-us/projects/objectclass recognition/} is the original version of the MSRC dataset. 
It includes 210 images, with 30 images available for each of the seven classes~\cite{winn2005locus}. 
Our experiment employs five different feature extractors: color moment, LBP, HOG, SIFT, and GIST.
\textbf{Outdoor-Scene}\footnote{https://pan.baidu.com/s/1A1J92aOf7wRY8Q\_7P5oidA Extracted code: ce13} comprises 2,688 color images, each classified into one of eight outdoor scene categories. 
We extract features from these images using four different visual features: color moment, GIST, HOG, and LBP~\cite{monadjemi2002experiments}.
\textbf{N-W-SCENE} is a subset of the NUS-WIDE dataset. It covers 33 scene concepts with a total of 34,926 images. 
This dataset contains five types of low-level features extracted from the images: color histogram, color correlogram, edge direction histogram, wavelet texture, and block-wise color moments~\cite{nus-wide-civr09}.
\textbf{N-W-OBJECT} is another subset of NUS-WIDE. It includes 31 object categories with 30,000 images for object-based tasks. The same five types of low-level features are extracted~\cite{nus-wide-civr09}.
\textbf{XRMB}\footnote{https://home.ttic.edu/~klivescu/XRMB\_data/full/README} (Wisconsin X-Ray Microbeam) is a dataset consisting of human dental X-ray images, specifically designed to study the impact of radiation on biological tissues. 
It includes two views: the acoustic view and the articulation view~\cite{wang2015unsupervised}.
The statistics for these multi-view datasets are presented in Table \ref{tab:datasets_statistics}.

\subsubsection{Comparative Methods} 
We compare our CUMI approach with several multi-view algorithms, including two baseline methods, two traditional multi-view algorithms, and three state-of-the-art methods, to demonstrate its effectiveness. The details are given as follows.

\begin{itemize}
    \item \textbf{CCA}~\cite{chaudhuri2009multi}: Serving as a baseline approach, CCA employs canonical correlation analysis to project high-dimensional data into a lower-dimensional subspace. Subsequently, it utilizes an SVM classifier for the classification task.
    \item \textbf{DCCA}~\cite{andrew2013deep}: DCCA uses deep neural networks (DNNs) to extract nonlinear features for each view. The method aims to maximize the canonical correlation between features extracted from different views. 
     \item \textbf{MVSS}~\cite{wang2013multi}: As a traditional method, MVSS applies joint group $\ell_1-$norm and $\ell_{2,1}-$norm regularization to achieve feature sparsity across selected views. This accounts for the varying discriminative information present in each view.
    \item \textbf{WeightReg}~\cite{yang2019adaptive}: WeightReg employs a regression-based structure and introduces a new discriminative regression target. This enhances the discrimination of features in the projected subspace while preserving their inherent characteristics.
    \item \textbf{DUA-Net}~\cite{geng2021uncertainty}: Recent state-of-the-art techniques have significantly improved the reliability and robustness of multi-view classification. DUA-Net employs a generative model to estimate data uncertainty, which in turn guides the learning process. This strategy dynamically integrates multiple data views by assigning weights according to their respective data uncertainties. 
    \item \textbf{MEIB}~\cite{zhang2022multi}: MEIB extends the information bottleneck principle to supervised multi-view learning, facilitating the fusion of complementary information across different views. 
    \item \textbf{TMC}~\cite{han2023trusted}: TMC leverages uncertainty estimation to dynamically assess the trustworthiness of each view, thereby enhancing the reliability and robustness of multi-view classification.
\end{itemize}

\begin{table}[t!]
\centering
\caption{Statistics of multi-view datasets.}
\label{tab:datasets_statistics}
\resizebox{0.5\textwidth}{!}{%
\begin{tabular}{@{}lllll@{}}
\toprule
Dataset         & Sample & class & view & dimensionality of features      \\ \midrule
Caltech101-7    & 1474   & 7       & 6    & \{48, 40, 254, 1984, 512, 928\} \\
Caltech101-20   & 2386   & 20      & 6    & \{48, 40, 254, 1984, 512, 928\} \\
MSRC         & 210    & 7       & 5    & \{24, 576, 512, 256, 254\}      \\
Outdoor-Scene   & 2688   & 8       & 4    & \{512, 432, 256, 48\}           \\
N-W-SCENE  & 34926  & 33      & 5    & \{64, 225, 144, 73, 128\}       \\
N-W-OBJECT & 30000  & 31      & 5    & \{64, 225, 144, 73, 128\}       \\
XRMB            & 7000   & 10      & 2    & \{273, 112\}                    \\ \bottomrule
\end{tabular}%
}
\end{table}

\subsubsection{Details of Implementation}
In our experiments, we utilize a Multi-layer Perceptron (MLP) with ReLU activation functions.
Each view is followed by two encoders and a single decoder.
For a given view with dimension $d$ and $n$ categories, the encoder $\phi_{\mathcal{C}}$ is structured with three layers, following the sequence $d$-$1.2d$-$0.5d$. 
The final layer of encoder $\phi_{\mathcal{C}}$ consists of $10n$ neurons.
The encoder $\phi_{\mathcal{U}}$ follows a similar structure, its last layer contains $5n$ neurons. 
The decoder $\psi$ is constructed with four layers, reflecting the aggregate sum of the two encoders, specifically in the sequence $15n$-$d$-$2.4d$-$d$. 
The resulting latent representation is obtained by concatenating the final layers of the two encoders for each view, which leads to a total of $(5v+10)n$ neurons, with $v$ representing the number of views.
This combined representation, denoted as $\mathcal{Z}$, serves as the input for the classifier.
The parameters $\beta$ and $\gamma$ are fine-tuned within the set \{0.001, 0.01, 0.1\}.
The model is trained over 100 epochs using stochastic gradient descent (SGD) with a learning rate of 0.01.
For the methods used for comparison, we utilize open-source codes and adhere to the recommended settings provided by their respective authors.
For CCA and DCCA, we used the parameter settings provided by the authors in their code.
For MVSS, we optimize the parameters $\gamma_1$ and $\gamma_2$ within the range of \{$10^{-5}$, $10^{-4}$, ..., $10^4$, $10^5$\}.
Similarly, for WeightReg, we optimize the parameter $\gamma$ within the range of \{$10^{-3}$, $10^{-2}$, ..., $10^2$, $10^3$\}.
The DUA-Net model does not have explicit hyperparameters, and we follow the authors' suggestion of setting the latent representation dimension to 50.
For MEIB, we fine-tune the parameter $\beta$ within the set \{0.001, 0.01, 0.1\}.
Lastly, for TMC, we set the regularization parameter to $10^{-4}$.

\subsubsection{Evaluation of Performance}
The efficacy of the proposed model is assessed through four distinct metrics in our experiments: accuracy, precision, recall, and the F1-score. Each of these metrics captures different aspects of classification performance, yet they all share a universal attribute: a higher value signifies better classification results.
To attenuate the impact of random variation, each method is executed 10 times, and the mean performance is reported. 
This process ensures a more reliable and robust evaluation of the model's performance.

\subsubsection{Comparison with State-of-the-Arts}
Table~\ref{tab:performance} presents the performance of various multi-view methods for the classification task. 
\begin{table*}[ht!]
\centering
\caption{\textcolor{black}{Performance of multiview classification.}}
\label{tab:performance}
\resizebox{\textwidth}{!}{%
\begin{tabular}{cccccccccc}\toprule
Datasets                            & Metrics   & CCA        & DCCA       & MVSS       & WeightReg   & DUA-Net      & MEIB   & TMC        & CUMI    \\  \midrule
\multirow{4}{*}{\rotatebox{90}{Caltech101-7}}       & ACC       & 91.05±0.03 & 88.40±0.03 & 97.89±0.01 & \underline{98.58±0.01} & 91.79±1.76    & 98.50±0.84 & 98.17±0.92 & \textbf{98.64±0.74}   \\
                                    & Precision & 75.86±1.44 & 80.71±0.76 & 92.06±0.34 & 95.65±0.08  & 87.40±7.79               & \textbf{96.75±2.57} & 95.97±2.63 & \underline{96.72±2.24}   \\
                                    & Recall    & 64.83±0.72 & 50.18±0.57 & 89.08±0.37 & 91.76±0.17  & 65.08±6.27               & \underline{92.42±5.03} & 90.72±4.39 & \textbf{92.91±4.11}   \\
                                    & F1        & 74.71±0.13 & 73.92±0.42 & 89.52±0.34 & 92.22±0.16  & 71.38±6.54               & \underline{93.50±4.57} & 92.25±4.01 & \textbf{93.98±3.21}   \\ \cmidrule(r){1-10}
\multirow{4}{*}{\rotatebox{90}{Caltech101-20}}      & ACC       & 79.88±0.05 & 74.29±0.08 & 94.18±0.01 & \underline{94.35±0.02} & 83.19±1.92    & 94.25±0.01 & 91.28±1.50 & \textbf{94.72±1.42}   \\
                                    & Precision & 64.03±0.24 & 64.83±0.18 & 89.31±0.03 & \textbf{90.58±0.02}    & 85.05±6.80    & 89.01±0.04 & 88.31±5.05 & \underline{89.42±3.97}   \\
                                    & Recall    & 57.68±0.07 & 48.60±0.18 & 85.60±0.07 & \underline{86.17±0.10} & 60.87±4.73   & 85.07±0.04 & 77.49±4.23 & \textbf{86.46±4.09}   \\
                                    & F1        & 63.13±0.16 & 61.10±0.28 & 86.03±0.06 & \textbf{86.73±0.08}    & 66.17±4.97    & 85.21±0.04 & 80.05±4.66 & \underline{86.47±4.11}   \\ \cmidrule(r){1-10}
\multirow{4}{*}{\rotatebox{90}{MSRC}}            & ACC       & 62.86±1.65 & 40.48±1.17 & 81.90±1.25 & 80.48±1.64             & 79.05±7.13    & \underline{95.24±4.26} & 93.33±5.30 & \textbf{95.71±2.56}   \\
                                    & Precision & 68.90±1.47 & 44.17±1.38 & 84.98±1.07 & 82.90±1.49             & 82.95±7.18    & \underline{96.43±3.19} & 95.38±3.48 & \textbf{96.79±1.92}   \\
                                    & Recall    & 62.86±1.65 & 40.48±1.17 & 81.90±1.25 & 80.48±1.64             & 79.05±7.13    & \underline{95.24±4.26} & 93.33±5.30 & \textbf{95.71±2.56}   \\
                                    & F1        & 64.18±1.39 & 49.95±1.06 & 82.46±1.17 & 82.40±0.87             & 78.12±7.76    & \underline{95.10±4.38} & 93.06±5.48 & \textbf{95.59±2.64}   \\ \cmidrule(r){1-10}
\multirow{4}{*}{\rotatebox{90}{Outdoor Scene}}      & ACC       & 79.42±0.13 & 61.20±0.11 & 83.89±0.11 & 77.28±0.23             & 74.07±1.97    & 84.12±2.60 & \underline{84.49±1.94} & \textbf{89.88±1.72}   \\
                                    & Precision & 80.09±0.16 & 61.35±0.16 & 84.46±0.14 & 77.97±0.31             & 75.78±2.24    & 84.72±2.82 & \underline{85.59±1.79} & \textbf{90.18±1.95}   \\
                                    & Recall    & 79.60±0.17 & 60.90±0.11 & 84.22±0.14 & 77.67±0.32             & 75.07±2.06   & 84.30±2.53 & \underline{84.76±1.91} & \textbf{89.93±1.83}   \\
                                    & F1        & 79.44±0.18 & 60.52±0.13 & 84.10±0.14 & 77.46±0.32             & 74.02±1.98     & 84.33±2.60 & \underline{84.39±1.92} & \textbf{89.93±1.86}  \\ \cmidrule(r){1-10}
\multirow{4}{*}{\rotatebox{90}{N-W-SCENE}}     & ACC       & 40.81±0.00 & 40.76±0.00 & 33.74±0.01 & 35.37±0.01             & 35.52±0.83    & \underline{49.09±0.64} & 44.29±0.26 & \textbf{49.40±0.78} \\
                                    & Precision & \underline{28.82±0.84}    & \textbf{40.76±0.00} & 9.44±0.00 & 10.61±0.01  & 13.34±1.91            & 15.74±1.76 & 13.21±1.43 & 17.56±1.84 \\
                                    & Recall    & 3.12±0.00  &  3.09±0.00  & 8.94±0.00 & 9.76±0.01                          & 8.20±0.52            & \underline{10.60±0.67} & 5.98±0.29  & \textbf{12.08±0.91} \\
                                    & F1        & \underline{34.67±1.59} & \textbf{57.91±0.00} & 15.18±0.01 & 15.95±0.01    & 8.60±0.66            & 11.14±0.76 & 6.35±0.42  & 12.96±1.07 \\ \cmidrule(r){1-10}
\multirow{4}{*}{\rotatebox{90}{N-W-OBJECT}}    & ACC       & 25.67±0.01 & 24.60±0.01 & 33.14±0.02 & 33.86±0.01                         & 27.35±0.79            & \underline{45.21±0.81} & 38.24±0.58 & \textbf{48.51±0.85} \\
                                    & Precision & 20.38±0.04 & 21.24±0.09 & 24.92±0.03 & 26.05±0.03                         & 25.72±1.61            & 33.15±1.89 & \underline{34.93±2.83} & \textbf{42.58±1.83} \\
                                    & Recall    & 11.83±0.00 & 10.88±0.00 & 23.39±0.02 & 24.45±0.02                         & 19.04±0.78            & \underline{28.91±0.90} & 21.51±0.50 & \textbf{36.92±0.74} \\
                                    & F1        & 22.18±0.03 & 21.42±0.07 & 24.53±0.02 & 26.08±0.01                         & 19.20±0.79            & \underline{28.76±0.84} & 21.76±0.58 & \textbf{37.78±0.70} \\ \cmidrule(r){1-10}
\multirow{4}{*}{\rotatebox{90}{XRMB}}               & ACC       & 55.90±0.03 & 68.14±0.08 & 70.29±0.02 & 71.64±0.03                         & 67.20±1.65            & \underline{76.34±0.93} & 67.59±1.17 & \textbf{81.87±1.16} \\
                                    & Precision & 55.82±0.03 & 68.74±0.08 & 70.55±0.02 & 71.82±0.03                         & 68.82±1.90           & \underline{76.25±0.97} & 68.61±1.30 & \textbf{82.01±1.11} \\
                                    & Recall    & 55.90±0.03 & 68.14±0.08 & 70.29±0.02 & 71.64±0.03                         & 67.20±1.65            & \underline{76.34±0.93} & 67.59±1.17 & \textbf{81.87±1.16} \\
                                    & F1        & 55.52±0.04 & 68.16±0.08 & 70.27±0.02 & 71.57±0.03                         & 67.07±1.79           & \underline{76.15±0.94} & 66.19±1.33 & \textbf{81.85±1.15} \\  \bottomrule
\end{tabular}%
}
\end{table*}
Overall, our algorithm consensusly outperforms other comparison methods across all datasets.
As a baseline method, CCA-based algorithms do not manifest notable performance on the majority of datasets, specifically on the N-W-OBJECT, XRMB, and MSRC datasets. 
This underperformance can be attributed to the fact that CCA algorithms focus exclusively on linear correlations, thereby limiting their ability to cope with intricate relationships.
Despite DCCA algorithms' ability to effectively manage complex correlations, they uniformly map different views onto a projection subspace by maximizing the correlations. 
This approach leads to their inability to learn the complementary information from multiple views, which can consequently degrade the quality of the learned representation.
In contrast, our method exhibits significant improvements over recent methods such as MEIB and TMC. 
Our method effectively captures the unique information of each view and the common latent structure among different views, thereby enhancing the generalization capability of our model.
This clear superiority underscores the effectiveness of our proposed method.

\subsubsection{Statistical Test}
\begin{figure}[t]
    \centering
    \includegraphics[width=0.49\textwidth]{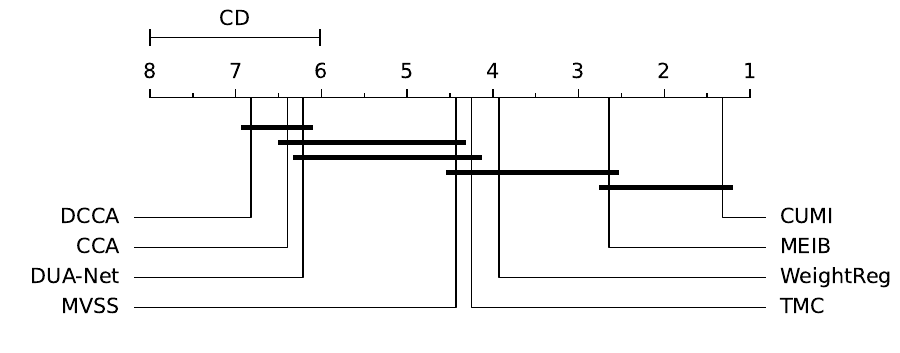}
    \caption{Results of the Nemenyi test indicating significant differences among the methods.
    The groups consisting of DCCA, CCA, and DUA-Net; CCA, DUA-Net, and MVSS; DUA-Net, MVSS, and TMC; MVSS, TMC, WeightReg, and MEIB; and MEIB and CUMI exhibit no significant differences.
    However, our method, CUMI, exhibits significant differences compared to the majority of the other methods.}
    \label{fig:Nemenyi_test}
\end{figure}

To ascertain the statistical significance of the experimental findings, we employed the Nemenyi test \cite{herbold2020autorank,demvsar2006statistical}. 
\textcolor{black}{The Nemenyi test is a non-parametric post-hoc analysis method utilized to compare multiple treatments or groups. It is particularly useful for identifying significant differences in the rankings of various groups' performances when the underlying assumptions for parametric tests are not met. This makes the Nemenyi test an appropriate choice for our analysis, where data may not follow a normal distribution.}
Based on the results depicted in Figure~\ref{fig:Nemenyi_test}, our method, CUMI, demonstrates a substantial advantage over the majority of the other methods. 
Figure~\ref{fig:Nemenyi_test} presents the Nemenyi test, where higher average ranks indicate superior performance (i.e., the right side is better than the left side). 
Methods connected by a solid horizontal line do not display statistically significant differences in mean ranks.
For reference, the critical distance, which indicates significant differences, is illustrated in the plot (CD = 1.984, $p$-value $<$ 0.05).
Upon analyzing Figure~\ref{fig:Nemenyi_test}, it is evident that there are no significant differences within the following groups: DCCA, CCA, and DUA-Net; CCA, DUA-Net, and MVSS; DUA-Net, MVSS, and TMC; MVSS, TMC, WeightReg, and MEIB; MEIB and CUMI. 
All other comparisons yield significant differences. 
\textcolor{black}{In other words}, with the exception of MEIB, our method demonstrates significant differences compared to all other methods.

\subsection{Analysis of Parameters and Model}
\begin{figure*}[t!]
    \centering
    \begin{subfigure}[b]{0.45\textwidth}
        \centering
        \includegraphics[width=\textwidth]{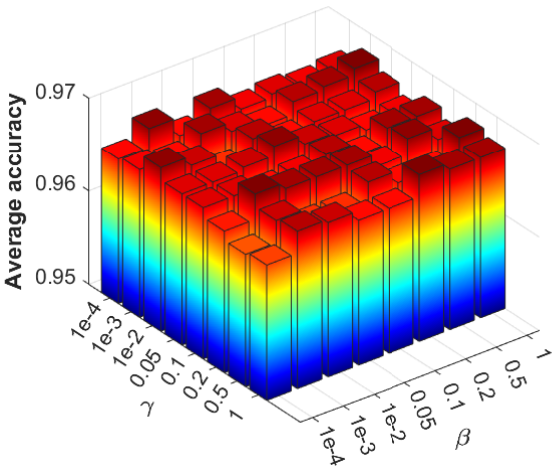}
        \caption{Parameter Analysis}
        \label{fig:PSA_Caltech101}
    \end{subfigure}
    \hfill
    \begin{subfigure}[b]{0.45\textwidth}
        \centering
        \includegraphics[width=\textwidth]{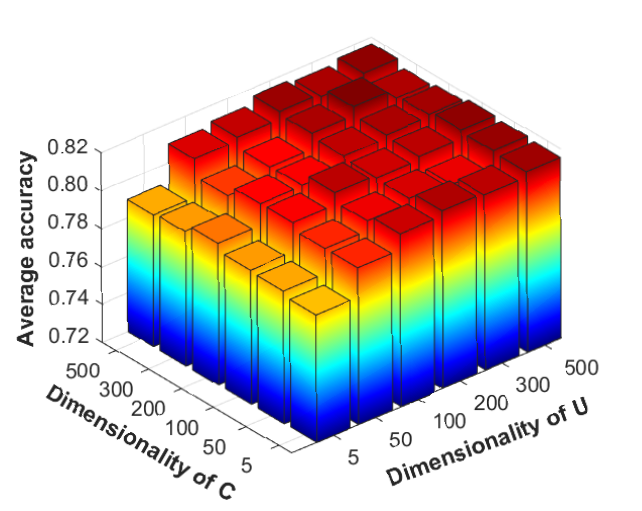}
        \caption{Model Analysis}
        \label{fig:MA}
    \end{subfigure}
    \caption{These two figures collectively present an analysis of the impact of hyperparameters ($\beta$ and $\gamma$) and model structure (common and unique feature dimensions) on the performance of the CUMI framework.
    }
    \label{fig:psa_ma}
\end{figure*}

This section delves into the influence of hyperparameters and the model structure on the performance of our proposed CUMI framework, as shown in Figure~\ref{fig:psa_ma}. 
\subsubsection{Sensitivity Analysis of Parameters}
We scrutinize the sensitivity of the model parameters using the Caltech101-7 dataset, with $\beta$ and $\gamma$ varying within the range of \textcolor{black}{$[10^{-4}$, $10^{-3}$, $10^{-2}$, 0.05, 0.1, 0.2, 0.5, 1]}. 
As illustrated in Figure~\ref{fig:PSA_Caltech101}, we adopt a grid search strategy to assess their average classification performance across five repeated experiments.

Intriguingly, our method exhibits a high degree of robustness, showing little sensitivity to the specific choices of $\beta$ and $\gamma$. This suggests that our multi-view learning framework effectively amalgamates information from different views, discriminates between common and unique information, and thereby enhances its resilience.

\subsubsection{Model Structure Analysis}
We explore the effects of the common feature dimension $\mathcal{C}$  and unique feature dimension $\mathcal{U}$ on the performance of our framework on XRMB dataset. Figure~\ref{fig:MA} shows the results of employing a grid search strategy to assess the average classification accuracy across five experiments, with the common feature dimension and unique feature dimension varied within the range of \{$5, 50, 100, 200, 300, 500$\}.
Figure~\ref{fig:MA} reveals that when the common and unique dimensions are insufficient, the performance falls short of expectations. This is intuitively understood as insufficient feature dimensions being unable to capture adequate task-relevant information. Boosting both dimensions enhances model performance. However, we note that beyond a certain threshold, further augmenting the feature dimensions does not yield performance improvement. This underscores the need to eschew excessive model complexity and select feature dimensions judiciously.

\subsubsection{Time Complexity Analysis}\label{sec:Time Complexity Analysis}
In this section, we will discuss the complexity of our algorithm. 
The term TC in Equation (\ref{eq:Renyi_TC_our}) involves eigenvalue decomposition, which can be computationally expensive. 
Therefore, we conducted comparative experiments by removing the TC term from our algorithm CUMI, denoted as CUMI/TC.
We tested both methods on the XRMB dataset for 100 epochs. 
Our CUMI algorithm took 101.6 seconds to run with an average accuracy of 81.81\%. 
On the other hand, CUMI/TC took 70.4 seconds with an average accuracy of 80.99\%, without the TC computation.

The results indicate that the time required for TC calculation is acceptable given the performance improvements it provides. 
However, we would like to emphasize that a recent work~\cite{dong2023optimal} presents computationally efficient approximations that can significantly reduce the complexity of R{'e}nyi's $\alpha$-order entropy functional to even less than $O(n^2)$. 
This implies that our running time can be substantially reduced with a theoretical guarantee. 
We plan to explore this possibility in future work.

\section{Conclusion and Future Work}
This paper presents a novel multi-view learning framework, underpinned by an innovative and mathematically rigorous definition of multi-view common information. 
Our approach is designed to capture both common and unique information from each view, integrating them into a unified representation through the minimization of a total correlation term. This process ensures the independence of the extracted common and unique information, providing a theoretical guarantee of our framework's effectiveness.
The novel aspect of our framework is that it circumvents the need for variational approximation and distributional estimation in high-dimensional space. 
Instead, it employs a matrix-based R{\`e}nyi's $\alpha$-order entropy functional to estimate common information. 
Experimental evaluations on real-world datasets underscore the superior performance of our proposed framework over existing methods, as it adeptly recognizes and utilizes both common and unique information across diverse views.

The primary limitation of our proposed method is the lengthy training duration due to the computation of the matrix-based R{\`e}nyi's $\alpha$-order entropy functional, as discussed in Section \ref{sec:Time Complexity Analysis}.
Future research aims to develop efficient approximations for this entropy functional, reducing the complexity of our CUMI framework.
Furthermore, we will also investigate the application of our method in scenarios with incomplete multiview data, taking into account the definition of common information.

\section*{Acknowledgments}
This work was supported by the National Natural Science Foundation of China under grant number U21A20485, 62088102 and 62311540022.



\bibliographystyle{elsarticle-num} 
\bibliography{mybib.bib}





\end{document}